\documentclass{article}
\usepackage{url}
\usepackage{microtype}
\usepackage{parskip}
\usepackage[super]{natbib}
\usepackage[a4paper, left=2.5cm, right=2.5cm, top=2.5cm, bottom=2.5cm]{geometry}
\usepackage{longtable,booktabs}
\usepackage{caption}
\usepackage{blindtext}
\usepackage{graphicx}
\usepackage{authblk}
\usepackage{amsmath}
\usepackage{lineno}
\usepackage[toc,page]{appendix}
\usepackage[utf8]{inputenc}

\usepackage[ruled,vlined]{algorithm2e}

\title{Developing an OpenAI Gym-compatible framework and simulation environment for testing Deep Reinforcement Learning agents solving the Ambulance Location Problem\\}

\author[1]{*Michael Allen}
\author[1]{Kerry Pearn}
\author[2]{Thomas Monks}

\affil[1,*]{\footnotesize University of Exeter Medical School \& NIHR South West Peninsula Applied Research Collaboration (ARC).}
\affil[2]{\footnotesize University of Exeter Institute of Data Science and Artificial Intelligence}
\affil[*]{\footnotesize Corresponding author: m.allen@exeter.ac.uk}

\begin{document}

\maketitle

\section*{Abstract} 

\emph{Background and motivation}: Deep Reinforcement Learning  (Deep RL) is a rapidly developing field. Historically most application has been made to games (such as chess, Atari games, and go). Deep RL is now reaching the stage where it may offer value in real world problems, including optimisation of healthcare systems. One such problem is where to locate ambulances between calls in order to minimise time from emergency call to ambulance on-scene. This is known as the Ambulance Location problem.

\emph{Aim}: To develop an OpenAI Gym-compatible framework and simulation environment for testing Deep RL agents.

\emph{Methods}: A custom ambulance dispatch simulation environment was developed using OpenAI Gym and SimPy. Deep RL agents were built using PyTorch. The environment is a simplification of the real world, but allows control over the number of clusters of incident locations, number of possible dispatch locations, number of hospitals, and creating incidents that occur at different locations throughout each day.

\emph{Results}: A range of Deep RL agents based on Deep Q networks were tested in this custom environment. All reduced time to respond to emergency calls compared with random allocation to dispatch points. Bagging Noisy Duelling Deep Q networks gave the most consistence performance. All methods had a tendency to lose performance if trained for too long, and so agents were saved at their optimal performance (and tested on independent simulation runs).

\emph{Conclusion}: Deep RL agents, developed using simulated environments, have the potential to offer a novel approach to optimise the Ambulance Location problem. Creating open simulation environments should allow more rapid progress in this field.

\emph{GitHub repository of code}: \url{https://github.com/MichaelAllen1966/qambo}

\section*{Keywords}
Health Services Research, Health Systems, Simulation, Reinforcement Learning, Ambulance

\section{Introduction}

Deep Reinforcement Learning (Deep RL) and Health System simulations are two complementary and parallel methods that have the potential to improve the delivery of health systems.

Deep RL is a rapidly developing area of research, finding application in areas as diverse as game playing, robotics, natural language processing, computer vision, and systems control \cite{li_deep_2018}. Deep RL involves an \emph{agent} that interacts with an \emph{environment} with the aim of developing a \emph{policy} that maximises long term \emph{return} of \emph{rewards}. Deep RL has a framework that allows for generic problem solving that is not dependent on pre-existing domain knowledge, making these techniques applicable to a wide range of problems.

We have focused on methods based on \emph{Deep Q Learning}. These methods predict the best long-term accumulated reward (\emph{Q}) for each action, and recommend the action with the greatest \emph{Q}. As knowledge of actual rewards are accumulated, the networks better predict \emph{Q} for each action. 

Health Systems simulation seeks to mimic the behaviour of real systems in a simulated environment. These may be used to optimise services such as emergency departments \cite{monks_using_2017}, hospital ward operation and capacity \cite{penn_towards_2020} and ambulance handover at hospitals \cite{clarey_ambulance_2014}. These examples of health service simulations are used for off-line planning and optimization of service configuration. 
OpenAI Gym (\url{gym.openai.com}) is a standardised environment structure and Python library for developing and testing of Deep RL agents. We have previously demonstrated that the established Python Discrete Event Simulation library,  SimPy \cite{team_simpy_simpy_2020}, may be used to create simulated healthcare system environments compatible with OpenAI Gym, and have shown how Deep RL agents (based on Deep Q learning) may control staffed beds in a simplified hospital simulation environment \cite{allen_integrating_2020}.

The \emph{Ambulance Location Problem} is a classic problem in Healthcare Systems Operational Research \cite{brotcorne_ambulance_2003}. The problem is choosing which predefined dispatch location to locate free ambulances in order to minimise time from emergency call to ambulance on-scene. While this is a problem that may be expected to be amenable to reinforcement learning, a recent review found little evidence of application of Deep RL in the ambulance location problem \cite{tassone_comprehensive_2020}, though some work has very recently been published \cite{ji_deep_2019, liu_ambulance_2020}.

Our key aims were to 1) develop and pilot an open ambulance location simulation environment that would allow for testing of alternative Deep RL agents, and 2) to test some standard Deep RL agents. We aim to develop this ambulance location environment over time, with the ultimate aim of creating \emph{digital twins} of real environments that may be used for both research and for operational training of Deep RL agents.

\section{GitHub repository}

The GitHub repository containing this code, including a Jupyter notebook is provided for each type of Deep RL agent tested, may be found at: \url{https://github.com/MichaelAllen1966/qambo}

The examples cited in this paper are from release version 1.0.0 (DOI: 10.5281/zenodo.4432503): 
\url{https://github.com/MichaelAllen1966/qambo/releases/tag/v1.0.0}.

A Jupyter notebook is provided for each type of Deep RL agent tested.

\section{Method}

\subsection{Terminology}

Some key terminology used in this paper:

\begin{itemize}

    \item \emph{Simulation Environment}: The simulated behaviour of a system of emergency \emph{incidents} requiring conveyance of patients to {hospital} in \emph{ambulances}. The simulation includes a constrained \emph{world} of a given size. When free, ambulances wait at \emph{dispatch points} spread throughout the world.

    \item \emph{Agent}: The agent instructs which dispatch point to send an ambulance to when free. In this paper all agents, apart from one that randomly assigns dispatch points, are Deep RL agents based on Deep Q Learning.

    \item \emph{Gym}: OpenAI Gym (\url{gym.openai.com}) is a standardised environment structure and Python library for developing and testing of Deep RL agents.
        
    \item \emph{Model}: The 'model' refers to the combination of the agent and the simulation environment.

    \item \emph{SimPy}. A Python Discrete Event Simulation library.

\end{itemize}

\subsection{Simulation Environment}

The simulation environment is a simplification of the real world problem that allows control over the action of the objects contained. This section describes it in more detail.

\subsubsection{Model Overview}

\begin{itemize}
    
    \item Incidents occurs in areas within a world with fixed dimensions. The geographic pattern of incidents may change throughout the day.

    \item When an incident occurs, ambulances are dispatched from fixed dispatch points; the closest free ambulance is used.

    \item The ambulance collects a patient and conveys them to the closest hospital.

    \item The ambulance is then allocated, by the agent, to any dispatch point. The ambulance travels to that dispatch point where it becomes available to respond to incidents (the ambulance may also be allocated while travelling to a dispatch point, depending on simulation environment settings).

    \item The job of the agent is to allocate ambulances to dispatch points in order to minimise the time from incident to arrival of ambulance at the scene of the incident.

\end{itemize}

Algorithm \ref{algo:overview} shows a high level structure of the code. This will be common to all interactions of Deep RL agents and SimPy simulations with only Deep RL-specific alterations.

\begin{algorithm}[H]
\caption{High level view of the code (the Deep RL agent in this example is a Double Deep Q Network using a policy net, target net, and memory).}
\SetAlgoLined
Initialise simulation environment\;
Set up Deep RL agent (policy net, target net, memory)\;
\While{Training episodes not complete}{
    Reset sim\;
    \While{not in terminal state}{
        Get action from Deep RL agent policy net\;
        Pass action to simulation environment\;
        Take a step in simulation environment (until an ambulance requires allocation to a dispatch point)\;
        Agent receives (observations, reward, terminal, info) from simulation environment\;
        Add (observations, next state, reward, terminal) to memory\;
        Render environment (optional)\;
        Update policy net\;
        Save policy net if new best performance\;
        Update target net: copy policy weights to target every \emph{n} steps\;}
    }
Test performance of best policy net\;
\label{algo:overview}    
\end{algorithm}

\subsubsection{Simulation Environment Initiation}

The simulation environment is initiated just once for all of the training and testing of an agent. On initiation of the simulation environment object, the following are set up:

\begin{itemize}

    \item \emph{World coordinate system} with passed maximum \emph{x} and \emph{y} coordinates.

    \item \emph{Dispatch points} using the passed random seed.
    
    \item \emph{hospital locations points} using the passed random seed.
    
    \item \emph{Incident point centres} using the passed random seed. A different set of incident points is set for each epoch during the day.
    
\end{itemize}

\subsubsection{Simulation Environment Reset}

The environment is reset for each training and test run. On reset, the following are set up:

\begin{itemize}

    \item \emph{Ambulances}: These are all free at the beginning of the simulation, and are allocated randomly to dispatch points (each run will start with a different random allocation) and start the simulation at that dispatch point. One ambulance, however, is made available to be re-allocated - this is to align with the environment \emph{step} process which allocates a single ambulance each time (after this reset method, this only occurs after a patient is conveyed to hospital and the ambulance is allocated to a dispatch point).
    
    \item \emph{Incident process}: Incidents requiring an ambulance occur at random (or pseudo-random), with time between incidents sampled from a negative exponential distribution. Each run will have different random patterns. Each incident is added to an ordered list of unassigned incidents (those without an ambulance yet assigned). Each minute in the simulated world, this list is checked and ambulances assigned where possible using simple \emph{first-in-first-out} prioritisation.
    
    \item \emph{Observations}: The first set of observations is returned from the \emph{reset} method, with subsequent ones returned whenever a free ambulance requires an allocation to a dispatch point. See section \ref{obs} for details on observations.
    
\end{itemize}

\subsubsection{Simulation Step}
\label{obs}

The simulation environment step starts with an ambulance waiting to be allocated to a dispatch point. The step method contains the following key components:

\emph{Action}

The action passes the index of the dispatch point that will be allocated to the ambulance waiting for allocation (the ambulance has a dispatch point allocated each time they arrive with a patient at hospital). The ambulance then travels to that dispatch point at the speed given in the ambulance parameters (depending on the simulation settings, the journey may be interrupted to pick up a new patient if they are the closest ambulance at the time an ambulance is required, with the location of the ambulance calculated assuming straight line travel between the hospital and the assigned dispatch point).

\emph{Simulation time step loop}

The simulation proceeds in time steps of 1 minute. During that time ambulances may be travelling to hospital or to assigned dispatch points, new incidents may occur, and ambulances may be dispatched to incidents. The simulation breaks out of this loop if there is an ambulance that is waiting to be assigned a dispatch point (after conveying a patient to hospital).

\emph{Return of data to the agent}

\label{obs}

At the end of the simulation time step loop, when an ambulance is ready to be allocated to a dispatch point, the simulation environment returns observations, reward, terminal state, and info to the agent.

\begin{itemize}

    \item \emph{Observations:}  Observations contain data to describe the current state of the simulation environment. Observations are passed from the simulation environment to the agent on initiation of the simulation environment and with each model step, and returned as an one-dimensional array, the length of which is equal to the number of dispatch points + three. The first part of the observation is an array that is equal to the number of dispatch points, and is the number of ambulances currently allocated to each dispatch point (these may be ambulances present at the dispatch point, or may be ambulances currently travelling to that dispatch point). This is followed by the \emph{x} and \emph{y} coordinates of the position of the ambulance that the agent must next allocate a dispatch point to (this will be the \emph{x} and \emph{y} co-ordinates of the hospital that the ambulance has taken a patient to). The final element of the observation array is the time of day expressed as a fraction between 0 and 1.
    
    \item \emph{Reward:} Each time a patient is conveyed to hospital a reward is returned to the agent. The reward is the negative square of the time taken from call to ambulance arrival at scene of incident for that patient.
    
    \item \emph{Terminal:} whether the simulation has reached a determined maximum run time, in which case \emph{terminal = True}.
    
    \item \emph{Info:} The info dictionary contains all times for \emph{call-to-arrival} (time between patient call and ambulance arriving on scene), \emph{assignment-to-arrival} (time between an ambulance being assigned to an incident and arriving on scene), the total number of calls made so far, and the fraction of demand met so far (the number of calls where an ambulance has arrived on scene).
    
\end{itemize}

\subsubsection{Agent-simulation Environment Interface Methods}

In summary, there are three methods that interface the simulation environment and the agent:

\begin{itemize}

    \item \emph{reset:} resets the simulation environment to a starting state and passes the first set of observations to the agent.
    
    \item \emph{step:} the simulation steps between times when an ambulance is waiting to be allocated to a dispatch point. The simulation environment returns a tuple of \emph{observations, reward, terminal, info}.
    
    \item \emph{render:} displays the current state of the simulation (optional).

\end{itemize}

\subsubsection{Baseline Simulation Environment Parameters}
\label{baseline}

For the results presented and discussed in this paper the simulation environment was set up with the following characteristics:

\begin{itemize}
    
    \item Size of world is 50km\textsuperscript{2}.
    
    \item One hospital is located at the centre of the world.
    
    \item Ambulances, on average, each respond to eight incidents per day each (a low utilisation is used so that call-to-response time is mostly dependent on placement on ambulances, rather than any queuing for ambulances in the system).
    
    \item Ambulances must arrive at a dispatch point before being available for incidents.
    
    \item Ambulances travel in straight lines at 60 kph.
    
    \item There are 25 dispatch points spaced evenly across the 50km\textsuperscript{2} world.
    
    \item There are two patterns of incident locations per day.
    
    \item Incidents occur with a random jitter of $\pm$ 2km in \emph{x} and \emph{y} around incident location centre.
    
    \item Three scenarios are tested: 1) one incident area at any time of day, and three ambulances, 2) two incident areas at any time of day, and six ambulances, and 3) three incident areas at any time of day, and nine ambulances. 
    
\end{itemize}

\subsubsection{Training and Testing}

Each agent was trained by 50 one-year simulated time periods. The first 10 years used entirely random action selection. After that agents switched to either epsilon-greedy exploration (a declining probability of choosing actions at random rather than taking the neural network recommended action), or used noisy and/or bagging networks to aid exploration (see below). At the end of the simulated run time, the environment passes back \emph{Terminal=True}. The best performing agent (judged by the maximum total reward) was saved for testing.

Each agent was tested in 30 independent one-year model runs.

\subsection{Deep Reinforcement Learning Agents}

Various Deep RL agents (based on Deep Q networks) were built using PyTorch and compared with the random acting agent. All agents had their performance tested in the custom environment as outlined in section \ref{baseline}. The agents tested were as follows:

\begin{enumerate}

    \item \emph{Random assignment}: Dispatch points are selected at random.
    
    \item \emph{Double Deep Q Network} (ddqn): Standard Deep Q Network, with policy and target networks \cite{van_hasselt_deep_2015}.
    
    \item \emph{Duelling Deep Q Network} (3dqn): Policy and target networks calculate Q from sum of *value* of state and *advantage* of each action (*advantage* represents the added value of an action compared to the mean value of all actions) \cite{wang_dueling_2016}. 
    
    \item \emph{Noisy Duelling Deep Q Network} (noisy 3dqn). Networks have layers that add Gaussian noise to aid exploration \cite{schaul_prioritized_2016}.
    
    \item \emph{Prioritised Replay Duelling Deep Q Network} (pr 3dqn). When training the policy network, steps are sampled from the memory using a method that prioritises steps where the network had the greatest error in predicting Q \cite{schaul_prioritized_2016}.
    
    \item \emph{Prioritised Replay Noisy Duelling Deep Q Network} (pr noisy 3dqn). Combining prioritised replay with noisy layers.
    
    \item \emph{Bagging Deep Q Network} (bagging ddqn), with 5 networks. Multiple networks are trained from different bootstrap samples from the memory \cite{osband_deep_2016}. Action may be sampled at random from networks, or a majority vote used.
    
    \item \emph{Bagging Duelling Deep Q Network} (bagging 3dqn), with 5 networks. Combining the bagging multi-network approach with the duelling architecture.
    
    \item \emph{Bagging Noisy Duelling Deep Q Network} (bagging noisy 3dqn) with 5 networks. Combining the bagging multi-network approach with the duelling architecture and noisy layers.
    
    \item \emph{Bagging Prioritised Replay Noisy Duelling Deep Q Network} (bagging pr noisy 3dqn) with 5 networks. Combining the bagging multi-network approach with the duelling architecture, noisy layers, and prioritised replay.

\end{enumerate}

A separate Jupyter Notebook is provided for each of these agents in the associated GitHub repository.

\section{Results}

Full results are given in the Jypyter Notebooks in the accompanying GitHub repository.

\begin{figure}
\centering
\includegraphics[width=0.9\textwidth]{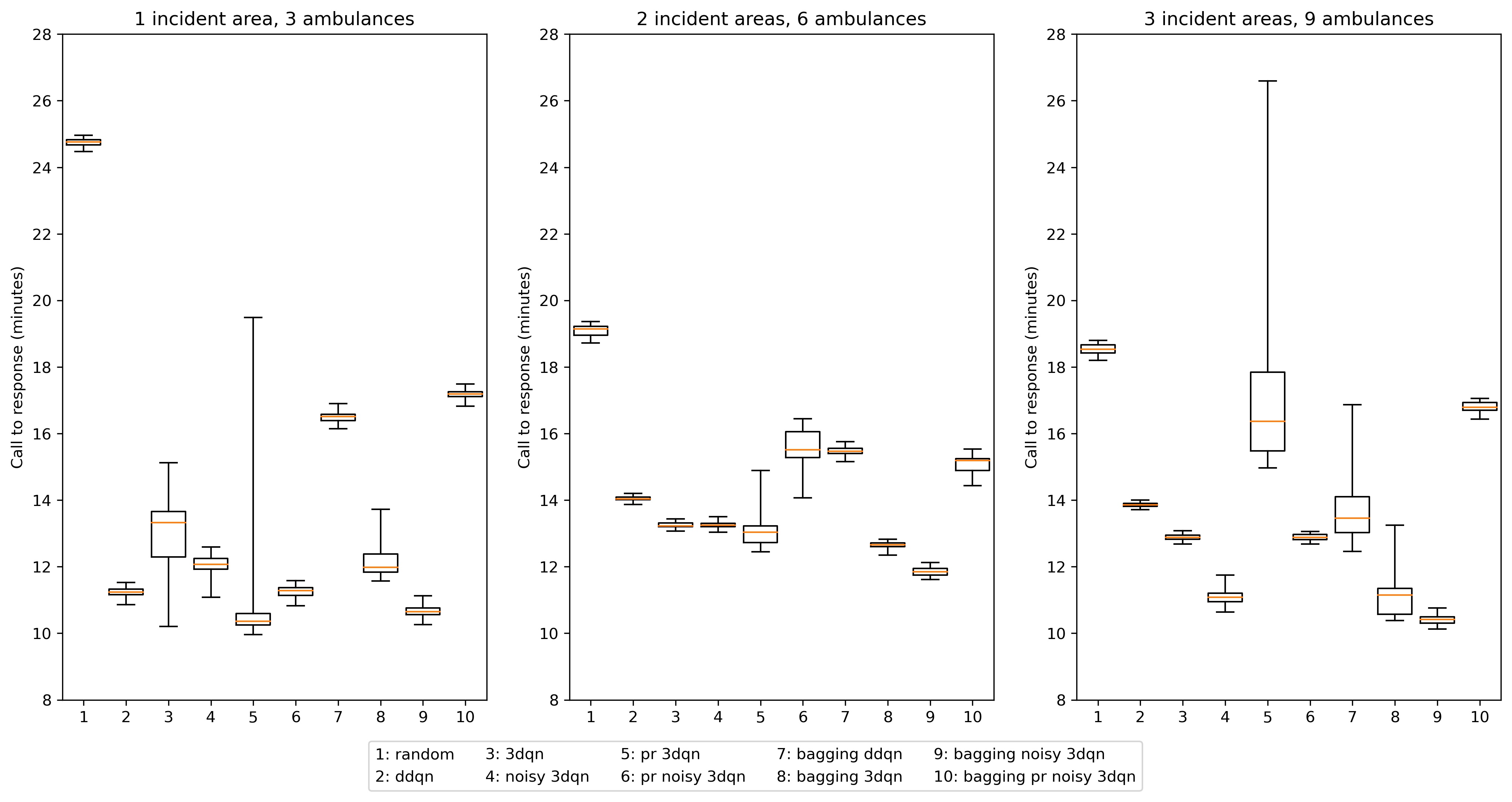}
\caption{Performance of a range of Deep Q RL agents across a range of scenarios. The scenarios are (left) one incident area at any time of day, and three ambulances, (centre) two incident areas at any time of day, and six ambulances, and (right) three incident areas at any time of day, and nine ambulances. All scenarios have two geographic patterns of incidents per day, and all have a single hospital. The ratio of ambulances to incidents is the same in all scenarios. The agents shown are 1) random assignment, 2) Deep Q Network (ddqn), 3) Duelling Deep Q Network (3dqn), 4) Noisy Duelling Deep Q Network (noisy 3dqn), 5) Prioritised Replay Duelling Deep Q Network (pr 3dqn), 6) Prioritised Replay Noisy Duelling Deep Q Network (pr noisy 3dqn), 7) Bagging Deep Q Network (bagging ddqn, with 5 networks), 8) Bagging Duelling Deep Q Network (bagging 3dqn, with 5 networks), 9) Bagging Noisy Duelling Deep Q Network (bagging noisy 3dqn, with 5 networks), 10) Bagging Prioritised Replay Noisy Duelling Deep Q Network (bagging pr noisy 3dqn with 5 networks). Boxes represent results from 30 test runs.}
\label{fig:boxplot}
\end{figure}

Figure \ref{fig:boxplot} shows results for different agents across three scenarios with increasing numbers of incident areas and ambulances. All agents improved performance compared to random allocation to dispatch points. The agent with the best performance in this test was the Bagging Noisy Duelling Deep Q Network, combining the advantages of duelling networks, noisy networks, and training multiple networks using the bagging technique.

\begin{figure}
\centering
\includegraphics[width=0.45\textwidth]{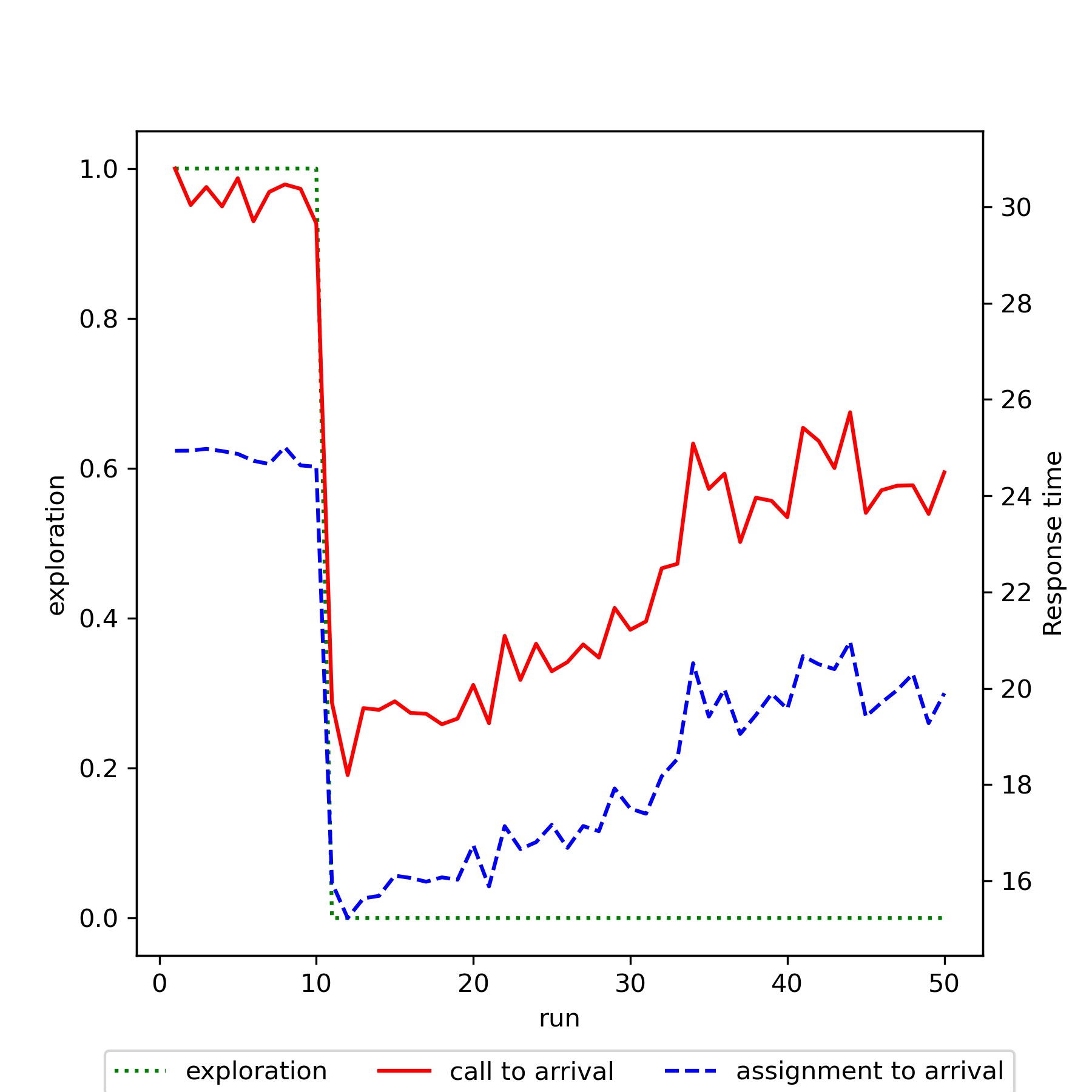}
\caption{Training of the Bagging Noisy Duelling Deep Q Network. The model acted purely randomly for 10 model runs and then switched to the Bagging Noisy Duelling Deep Q Network (5 networks) with no epsilon-greedy exploration. Chart shows epsilon-greedy exploration (green), mean time from ambulance call to arrival (red) and mean time from ambulance assignment to arrival (blue).}
\label{fig:training}
\end{figure}

A common observation across all agents is exemplified by the training of the Bagging Noisy Duelling Deep Q Network shown in figure \ref{fig:training}. After an initial period of completely random exploration, performance of the agent rapidly reaches an optimum. But with further training there is some reduction in performance of the agent, or increased instability of the performance. For this reason the agent network is saved at optimal performance (and tested in independent testing runs).

\section{Discussion}

We have presented a very simplified version of the 'real world' problem of where to locate free ambulances in order to minimise time to an ambulance arriving on the scene of an emergency. With this basic simulated environment we have shown the potential of Deep Reinforcement Learning to offer one solution to the Ambulance Location Problem, as explored and developed using an OpenAI Gym-compatible simulation environment.

All agents tested were able to out-perform random allocation of ambulances to dispatch points. But there appeared to be differences between the agents, with a Bagging Noisy Duelling Deep Reinforcement agent appearing best. Though not presented here, bagging agents also have the advantage of communicating uncertainty. As multiple agents are used with the bagging approach (with action either taken at random from multiple agents, or from a majority vote), higher certainty may be observed when all agents recommend the same action, and lower certainty may be observed when there is significant variation in recommended action from the different networks. Bagging does however come with higher computational expense as multiple agents are trained. More work is required to confirm the differences in performance of these agents.

An observation across all agents was the degradation of performance with prolonged training. This is especially relevant to the challenge of moving from simulated to real environments. One potential solution could be to use a digital twin of the real environment (which can replay all location and timings of incidents) for all training, and use the resulting agent in the real world, updating the agent only by transfer of new data to the digital twin for further training. Alternatively it may be possible to optimise agent performance to increase stability allowing for inline training ('live' training) in the real world.

\subsection{Further Work}

We have described only early pilot work. There are key areas for further work, which include:

\begin{itemize}

    \item Confirm performance of agents in broader tests.
    
    \item Test other Deep RL agents.
    
    \item Integrate with benchmark AI agents such as OpenAI baselines (\url{https://github.com/openai/baselines}) and TensorFlow agents (\url{https://github.com/tensorflow/agents}). The use of a Gym-based simulation environment should allow easy use of these optimised agents.
    
    \item Compare performance of agents with more classical Operation Research techniques.
    
    \item Progressively increase the complexity of the simulated environment towards becoming a digital twin of a real world simulation. Ambulance organisations collect detailed data on time and locations of incidents, as well as associated data (such as proportion of incidents dealt with on-scene, time on scene, travel times, and times taken at hospital to transfer the patient to the care of the hospital). 
    
\end{itemize}

\section{Conclusion}

With this pilot, we show the feasibility of modelling ambulance dispatch location assignment, and the potential of Deep RL to find good solutions.

\section{References}
\bibliographystyle{ieeetr}
\bibliography{refs}

\section*{Funding}

This study was funded by the National Institute for Health Research (NIHR) Applied Research Collaboration (ARC) South West Peninsula. The views and opinions expressed in this paper are those of the authors, and not necessarily those of the NHS, the National Institute for Health Research, or the Department of Health.

\end{document}